\documentclass{archive}

\usepackage[style=apa,natbib=true]{biblatex}
\title{The Stacked Autoencoder Evolution Hypothesis}

\auth{
    Hiroyuki Iizuka \affil{1}
}
\corresponding{Hiroyuki Iizuka}{iizuka@chain.hokudai.ac.jp}

\affiliations{
    \item Center for Human nature, AI, and Neuroscience (CHAIN), Hokkaido University,  \\
    North 12, West 7, Kita-ku, Sapporo, Hokkaido, JAPAN    
}

\abs{
This study introduces a novel theoretical framework, the Stacked Autoencoder Evolution Hypothesis, which proposes that biological evolutionary systems operate through multi-layered self-encoding and decoding processes, analogous to stacked autoencoders in deep learning. Rather than viewing evolution solely as gradual changes driven by mutation and selection, this hypothesis suggests that self-replication inherently compresses and reconstructs genetic information across hierarchical layers of abstraction. 
This layered structure enables evolutionary systems to explore diverse possibilities not only at the sequence level but also across progressively more abstract layers of representation, making it possible for even simple mutations to navigate these higher-order spaces.
Such a mechanism may explain punctuated evolutionary patterns and changes that can appear as if they are goal-directed in natural evolution, by allowing mutations at deeper latent layers to trigger sudden, large-scale phenotypic shifts. To illustrate the plausibility of this mechanism, artificial chemistry simulations were conducted, demonstrating the spontaneous emergence of hierarchical autoencoder structures. This framework offers a new perspective on the informational dynamics underlying both continuous and discontinuous evolutionary change.
}

\keywords{evolution, autoencoder, hypothesis}

\begin{document}

\coverpage           %
\doublespacing       %


\section{Introduction}
Since the foundational work of Darwin, evolutionary theory has been developed primarily on the basis of the principles of gradual change through adaptation and natural selection \citep{darwin1859}. According to this perspective, evolution proceeds not through sudden, large-scale transformations, but through the accumulation of small variations over extended periods, eventually giving rise to new traits and species. This model of gradual evolution has provided a powerful framework for explaining the diversity and complexity of life.

However, evidence from fossil records and molecular phylogenetic analyses has revealed that evolutionary change does not always proceed as a smooth, continuous process. Instead, it often alternates between long periods of stasis and episodes of rapid change. In particular, a number of well-documented evolutionary events exhibit marked discontinuities that challenge explanations based solely on incremental low-level mutations. Examples include the rapid morphological diversification during the Cambrian explosion \citep{mcmenamin2013cambrian}, the emergence of major body plans \citep{valentine2004origin}, and large adaptive radiations such as Darwin’s finches or African cichlids \citep{schluter2000ecology}. These episodes are characterized by sudden expansions of morphological or functional diversity, suggesting coordinated shifts across many traits rather than gradual accumulation of independent changes. These macroevolutionary discontinuities are not limited to the fossil record. Research in evolutionary developmental biology (Evo-Devo) has likewise shown that reorganisations of gene-regulatory networks or shifts in developmental constraints can produce coordinated, large-scale phenotypic transitions \citep{carroll2005endless}.Such discontinuities have played a central role in shaping present-day biodiversity by rapidly opening new regions of morphospace and enabling transitions that are difficult to achieve through purely gradualist mechanisms.

This pattern is exemplified by the theory of punctuated equilibrium, which suggests that species undergo long-term stability interrupted by brief periods of significant transformation \citep{eldredge1972punctuated}. Similarly, the long-standing hypothesis of saltation, or evolutionary ``leaps,'' proposes that large-scale changes in traits may arise suddenly through mutations or chromosomal rearrangements \citep{goldschmidt1940}. These hypotheses shed light on the discontinuous nature of evolution, suggesting that evolutionary change is not always uniform or gradual.

Despite the attention these hypotheses have received, they remain largely descriptive and leave the underlying mechanisms of such discontinuous changes poorly understood. For example, while punctuated equilibrium effectively describes patterns observed in the fossil record, it does not clarify the evolutionary dynamics driving these shifts. Likewise, although saltation has been occasionally observed, the likelihood that large-scale mutations result in viable organisms or species is generally considered low, making it difficult to regard saltation as a primary driver of evolution.

One frequently cited candidate mechanism for such changes is Kimura’s neutral theory of molecular evolution \citep{kimura1979neutral}. This theory posits that most molecular-level changes are neutral with respect to fitness and thus accumulate over long periods without manifesting in phenotypic change. It has been suggested that, under certain conditions such as sudden environmental shifts or developmental reorganization, these hidden and accumulated mutations might be simultaneously expressed, resulting in rapid phenotypic changes. While this mechanism offers a plausible explanation for punctuated and saltational changes, it fundamentally remains rooted in the Darwinian framework of incremental variation, framing evolution as a passive and stochastic process driven by random mutations and selection. Consequently, it falls short of fully accounting for the apparent leaps or purposefulness sometimes observed in evolutionary transitions.

In contrast, a more active view of evolution has been proposed by Imanishi \citep{asquith2007sources}. His theory, often referred to as Imanishi's evolutionary theory, portrays evolution not as a series of random events subject to external selection pressures, but as a process in which organisms actively choose their own transformations in pursuit of harmony with their environment. This perspective aims to account for evolutionary transitions that have sometimes been interpreted as exhibiting directionality or as if they were intentional in nature, without assuming any literal teleology.
However, the biological basis for such organismal agency and intentional change remains largely unspecified, and the levels at which such processes might operate have not been sufficiently clarified either theoretically or empirically. As such, existing theories, while recognizing discontinuity and directionality in evolution, have yet to provide a unified explanation of the dynamic mechanisms underlying these phenomena.

Addressing this gap, the present study proposes a novel perspective on evolutionary dynamics by focusing on the self-organizing exploratory capacity and informational dynamics inherent in evolutionary systems. In particular, it is hypothesized that the self-replication process itself may embody an information processing architecture analogous to stacked autoencoders, a concept that has advanced in the field of artificial intelligence. Through iterative self-replication, organisms may naturally compress and reconstruct genetic information, thereby forming multi-layered latent spaces within their ontogenetic architecture. Mutations occurring within these hierarchically structured latent spaces are expected to facilitate sudden and seemingly purposeful changes, providing a new explanatory framework for evolutionary leaps.

\section{Theoretical Background}
\subsection{Evolutionary processes in ALife and self-replicating systems}
The study of evolutionary processes has long been a central focus not only in biology but also in the field of artificial life. ALife research approaches the question of what constitutes life not by reducing it to its material components, but by emphasizing the mechanisms by which living systems sustain and generate their own organization and processes. Notably, the concept of autopoiesis \citep{varela1979}, which defines life as a system that continuously produces itself, along with models such as Langton’s self-replicating loop \citep{langton1984self}, which demonstrates self-replication based on minimal rule sets, has provided foundational frameworks for exploring the emergence and evolution of life-like behavior. Extending this line of research, EvoLoop has demonstrated that competition for spatial resources among self-replicating loops can give rise to evolutionary dynamics \citep{sayama1999new, salzberg2003emergent}. What is particularly important here is the idea that evolutionary processes can naturally emerge from material-level mechanisms of self-maintenance and self-replication, rather than being externally imposed or predefined.

These studies collectively highlight a perspective on evolution that moves beyond the conventional framing of evolution as a process of environmental optimization or goal-directed adaptation. Instead, they emphasize self-replication as a dynamic and ongoing process, through which systems continuously maintain and produce copies of themselves while remaining open to transformation and engagement with their environment. This perspective frames evolution not as a linear trajectory toward optimality, but as an open-ended process of recursive self-production and variation.

Experimental platforms such as Tierra \citep{ray1992} and Avida \citep{ofria2004avida} offer well-established computational environments for investigating evolutionary processes in artificial life. These systems simulate digital organisms, which are self-replicating programs, within virtual environments, enabling observation of how diversity and novel functions emerge through random mutations. Unlike earlier models such as autopoietic systems or self-replicating loops, these platforms operate at a slightly higher level of abstraction by embedding functional behavior directly into instruction sets. In particular, Avida has yielded detailed insights into how replication and accumulated mutations give rise to new behaviors. By reproducing cycles of self-replication and variation, these platforms have significantly deepened our understanding of evolutionary dynamics.

Although these platforms exhibit rich evolutionary dynamics and have demonstrated the emergence of novel functions, the genetic representations they employ are typically flat and linear. Their genomes are implemented as instruction sequences in which mutations modify individual instructions through local operations such as substitutions, insertions, or deletions. As a result, adaptive changes in these systems are generally modeled as arising from the gradual accumulation of small, local modifications to these sequences.

In contrast to instruction-sequence genomes, another line of ALife research has explicitly separated descriptions from executing operators. In the “machine–tape” models proposed by \citet{ikegami1995active}, self-reproduction is realized through networks of machines that read and write tapes, and evolutionary change arises not only from local perturbations of a fixed genome but also from the coupled dynamics between tapes (descriptions) and machines (operators). This framework highlights that replication can inherently involve an internal encode–decode cycle mediated by the interaction between representation and transformation mechanisms.

In addition to these simulation-based approaches, several analytical frameworks have sought to characterize the generative principles that underlie open-ended evolution. A notable example is the urn-based model of \citet{tria2014dynamics}, which formalizes Kauffman’s notion of the “adjacent possible” \citep{kauffman2000investigations}—the idea that the set of reachable novel states expands as new innovations accumulate. Such perspectives highlight that open-ended evolution is driven not only by variation and selection, but also by the dynamic enlargement of the possibility space itself. Situating the present hypothesis within this broader context suggests that hierarchical information-processing architectures, such as those proposed in this work, may provide an additional mechanism for generating and navigating these expanding evolutionary possibility spaces.

Building upon this body of research, the present study advances a novel hypothesis: that self-replication itself may inherently involve an internal process of self-encoding and decoding. Specifically, the self-replication process can be understood as a form of hierarchical information compression and reconstruction, analogous to the architecture of stacked autoencoders widely studied in machine learning. From this perspective, self-replication is not merely a process of copying existing structures, but rather a multi-layered information processing cycle in which organisms internally compress and reconstruct their structures before transmitting them to the next generation.

The following section provides a detailed explanation of stacked autoencoders, outlining their mechanisms.

\subsection{Stacked autoencoders: hierarchical feature learning in deep neural networks}
\begin{figure}[tp] 
    \centering
    \includegraphics[width=6.0in]{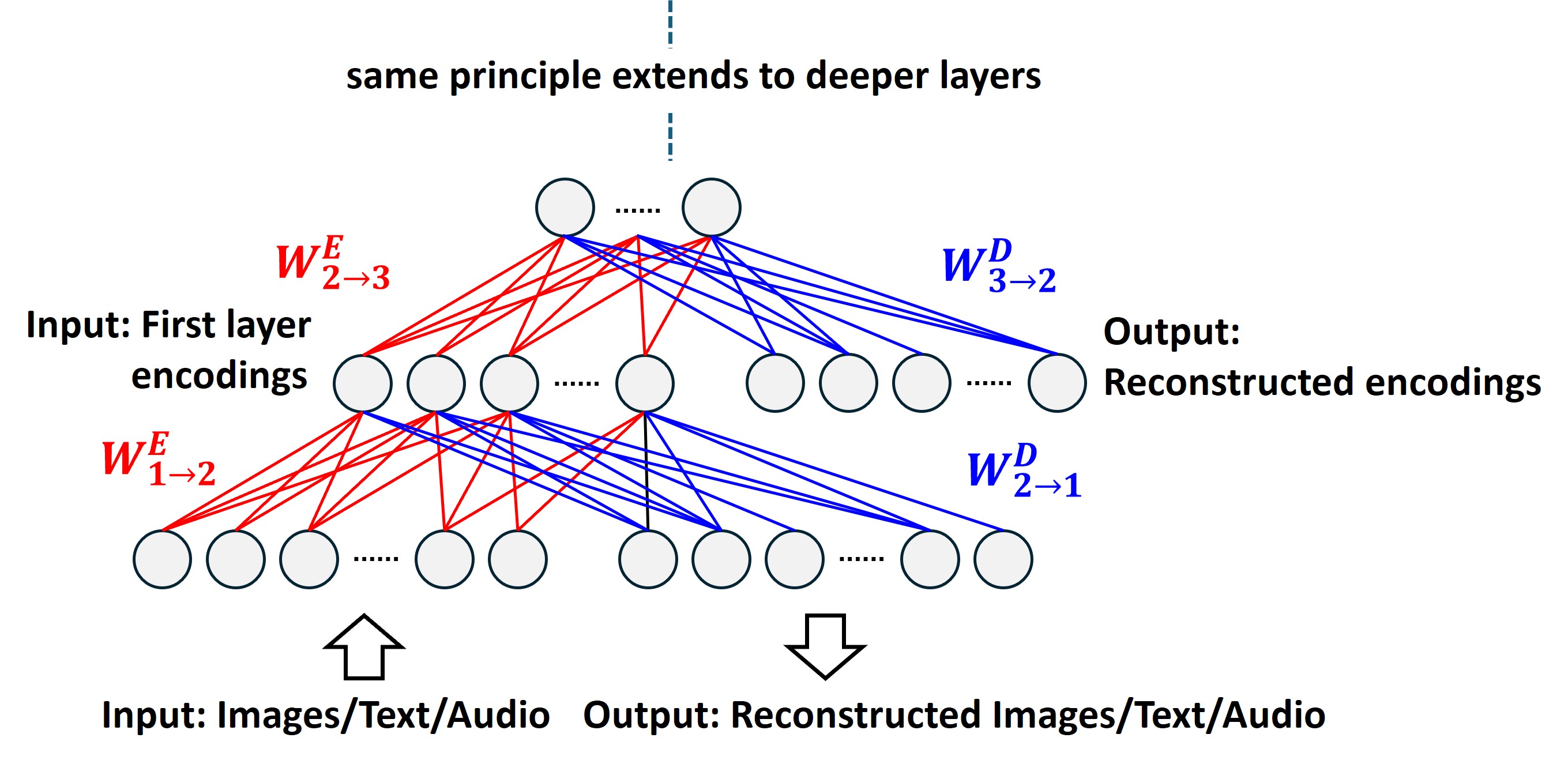}
    \caption{Schematic representation of a stacked autoencoder neural network. Red connections represent the encoding weights used to compress the input information, while blue connections represent the decoding weights used to reconstruct the input from the internal representations. Training is performed layer-wise.}
    \label{fig:sae}
\end{figure}

Stacked autoencoders represent a foundational architecture in deep learning, particularly for learning hierarchical feature representations through self-supervised learning \citep{bengio2006greedy, vincent2010stacked}. Traditionally, training deep neural networks suffered from issues such as vanishing gradients when layers were naively stacked. Stacked autoencoders address this by incrementally adding encoding and decoding layers, thereby enabling layer-wise pre-training and hierarchical abstraction. As shown in Fig.~\ref{fig:sae}, input data is progressively compressed into lower-dimensional latent spaces across layers, reducing the number of nodes and extracting increasingly abstract features \citep{bengio2013representation}. Shallow layers tend to capture low-level characteristics (e.g., edges, textures, character-level or word-level patterns), whereas deeper layers encode higher-level concepts such as object categories, semantic contexts, and abstract representations \citep{masci2011stacked, ngiam2011multimodal}. A key advantage of this architecture is that decoding layers are trained alongside encoding layers, allowing input reconstruction from intermediate feature representations. \citet{bengio2006greedy} demonstrated that stacked autoencoders mitigate the vanishing gradient problem and significantly improve training efficiency in deep networks, making them a milestone in the development of modern deep learning.

While stacked autoencoders employ deterministic autoencoding mechanisms, their conceptual basis in hierarchical representation learning parallels that of Deep Belief Networks (DBNs), which are based on probabilistic generative models \citep{hinton2006reducing, erhan2010does}. DBNs use restricted Boltzmann machines (RBMs) trained layer-by-layer, whereas stacked autoencoders rely on reconstruction objectives. Despite these differences, both architectures share the goal of discovering structured, multi-level representations. This study proposes that evolution may proceed through stacked autoencoder processes, a concept that is compatible with both deterministic autoencoders and probabilistic generative models such as DBNs.

\section{Stacked Autoencoder Evolution Hypothesis}
As suggested by theories such as punctuated equilibrium and saltation, evolutionary change is not always gradual but sometimes occurs in a punctuated or abrupt manner, albeit on different timescales. However, the underlying genetic mechanisms responsible for such rapid evolutionary transitions have yet to be fully elucidated. In this study, a novel evolutionary mechanism, Stacked Autoencoder Evolution Hypothesis, is proposed in order to address this gap from a different perspective.

\subsection{Hypothesis}
The central premise of the Stacked Autoencoder Evolution Hypothesis is that evolution is driven by a self-replication process based on self-encoding and decoding, analogous to the mechanism of an autoencoder. According to this view, evolution does not proceed merely through a simple iterative cycle of molecular replication, mutation, and selection. Instead, it advances through an informational process in which organisms self-replicate by internally maintaining and hierarchically reconstructing the genetic information embedded within their molecules across multiple layers of representation.

\begin{figure}[tp] 
    \centering
    \includegraphics[width=4.0in]{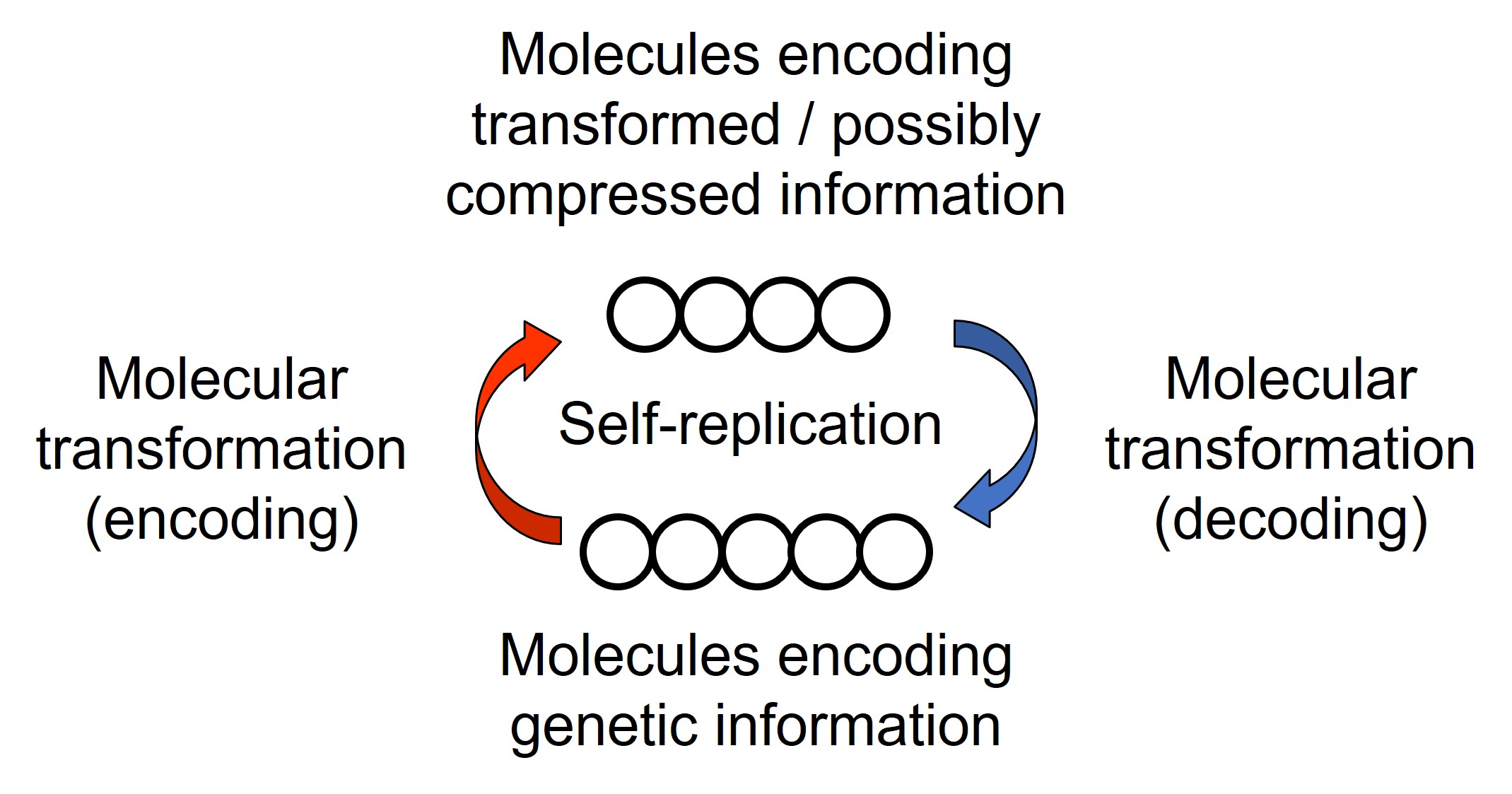}
    \caption{Self-replication conceptualized as a chain of molecular transformations. Intermediate molecules implement compression and reconstruction of genetic information, producing abstract latent representations analogous to autoencoder architectures.}
    \label{fig:autoenc}
\end{figure}

This hypothesis conceptualizes self-replication as a process in which genetic molecules undergo a chain of transformations facilitated by molecular interactions, ultimately reconstructing the original molecules (Fig.~\ref{fig:autoenc}). If these molecules are assumed to carry genetic information, then the molecular transformation process can be regarded as a transformation of that genetic information. In such cases, the information is represented within spaces of different dimensionality depending on the molecule that stores it. This is because the transformed molecule may be shorter than the original and capable of holding less information, or conversely, it may be longer. Crucially, even if the new molecule can only accommodate a smaller amount of information, the self-replication process must still find a way to preserve the original content, for example through compression. Such compression and reconstruction of information, realized through a chain of transformations in the self-replication process, correspond to the self-encoding and decoding operations seen in autoencoders. Through this process of compression and reconstruction, key features of the genetic information are extracted, resulting in increasingly abstract internal representations.

In the present formulation, we refer to “genes” and “genetic molecules” primarily for concreteness. However, the core requirement of the hypothesis is not tied to nucleotide-based genes per se, but to the existence of lower-level informational substrates that are explicitly encoded and participate in recursive transformation and self-replication cycles. In principle, such substrates could be instantiated by any structured units that are copied, transformed, and reconstructed across steps—for example, sequence-like regulatory modules or other compact encodings of developmental or behavioral patterns. In this paper, we focus on molecular-level genetic sequences as the most canonical example of such substrates, while leaving potential extensions to higher organisational levels for future work.

\begin{figure}[tp] 
    \centering
    \includegraphics[width=4.0in]{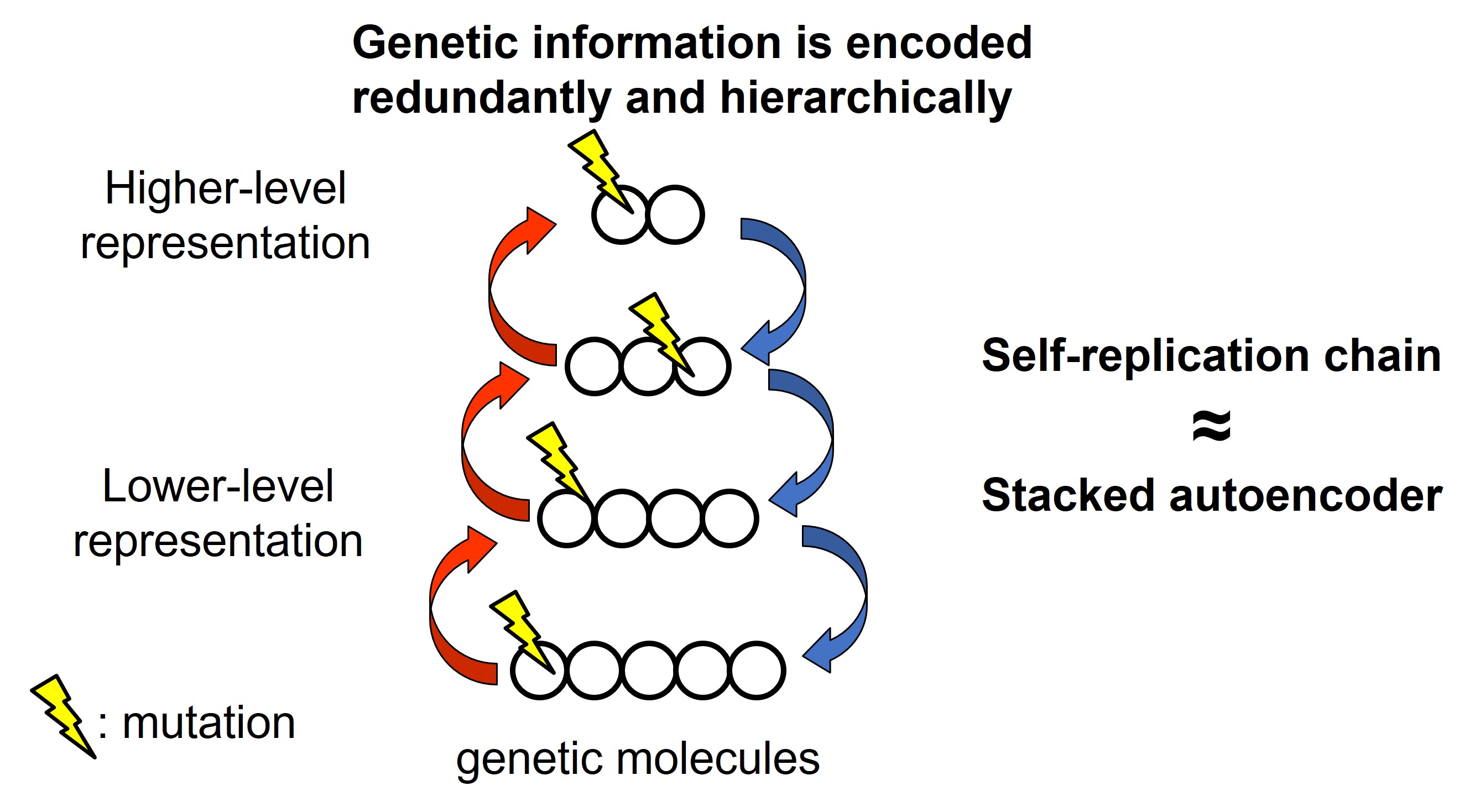}
    \caption{Mutations introduced at different layers of the replication hierarchy influence either physical sequences (shallow layers) or abstract latent representations (deep layers), with deeper perturbations being decoded into concrete molecular changes capable of producing discontinuous genetic novelty.}
    \label{fig:sae_hypo}
\end{figure}

Because the molecules with compressed information are themselves molecular entities, they may participate in the other self-replication processes (Fig.~\ref{fig:sae_hypo}). When replication proceeds via yet another molecule with an even smaller space, the information is further compressed. If this newly formed molecule also engages in a similar process, the self-replication chain continues, leading to successive stages of information compression. This recursive replication pathway mirrors the architecture of stacked autoencoders in deep learning, whereby genetic information is encoded into abstract spaces at successive levels, each characterized by increasing degrees of abstraction. Importantly, because each layer is physically instantiated by a distinct molecule, mutations can in principle occur at any level of this hierarchy—whether at the surface-level sequence itself or within deeper, more abstract latent spaces.

In this hierarchical structure, genetic information is redundantly maintained across multiple layers, with the deeper layers corresponding to higher-order latent spaces. Mutations introduced at shallow layers manifest as conventional physicochemical perturbations to the genetic sequence, whereas mutations occurring in deeper layers act as perturbations within an abstract feature space (Fig.~\ref{fig:sae_hypo}). These deeper perturbations need not correspond to continuous physical changes but instead shift the system within a higher-order representational manifold, potentially enabling qualitatively distinct, discontinuous evolutionary innovations. Through the downstream reconstruction processes of the replication cycle, these abstract mutations are eventually decoded back into concrete molecular sequences, thereby becoming physically realized and subject to natural selection. This mechanism provides a surprisingly direct route by which mutations in an abstract latent space can give rise to concrete genetic novelty.

As for how such encoding and decoding transformations can be acquired, the mechanism differs fundamentally from that used in artificial neural networks. In contrast to artificial neural networks, where hierarchical structures are typically acquired through layer-by-layer self-supervised training, evolution constructs these structures through self-organizing processes guided by selection pressure. As demonstrated in previous ALife simulations (e.g., \citep{ray1992,sayama1999new,ofria2004avida,hickinbotham2016maximizing}), molecular transformation processes that do not contribute to self-replication tend to be eliminated over time, whereas molecules that participate in effective self-replicating loops are naturally preserved. Although this evolutionary search mechanism may appear inefficient compared to the direct optimization strategies employed in artificial intelligence, it remains a viable process on evolutionary timescales. 

This hypothesis proposes that evolutionary systems inherently operate through a self-replication process that involves mechanisms analogous to stacked autoencoders, enabling efficient feature compression and reconstruction across multiple layers. The claim is not that literal neural-network–like modules exist within cells, but rather that an equivalent multi-layer encoding–decoding organisation could, in principle, emerge whenever multiple levels of structure are expressed within a shared molecular representation. In this sense, the architecture is conceptual, yet it points to a concrete class of implementable molecular organisations rather than to a purely metaphorical analogy.
As a result, this mechanism offers a potential explanation for how discontinuous and large-scale evolutionary transitions—sometimes appearing to follow directional or goal-like patterns—may arise without invoking actual teleology.

Although the present proposal is primarily conceptual, the hypothesis is in principle empirically testable. The stacked-autoencoder interpretation predicts that self-replication should involve intermediate, compressed representations that retain sufficient information to reconstruct higher-level structures. This prediction could be evaluated in several ways: for example, by analyzing whether perturbations to putative intermediate representations yield coherent, nonlocal changes in reconstructed structures; by examining synthetic or artificial-chemical systems for evidence of multi-stage information transformations; or by comparing the statistical geometry of reconstructed molecular states with that expected from hierarchical encoding models. These directions provide possible routes toward empirical or computational verification, even if direct biological tests remain challenging at present.

\subsection{Conceptual illustration of the hypothesis using artificial chemistry}
This section provides a concrete illustration of the proposed Stacked Autoencoder Evolution Hypothesis using the framework of artificial chemistry. Artificial chemistry is an abstract computational system that investigates self-replication and evolutionary processes through reaction networks composed of symbolic molecules. While these models do not directly incorporate ecological interactions or population-level dynamics, which are important for explaining biological evolution in full, they are well suited for describing the molecular transformation and autoencoder mechanisms that form the core of the present hypothesis. Integrating ecological and selective processes into this framework would likely provide a richer and more comprehensive understanding of evolutionary dynamics.

In this model, drawing inspiration from the RNA world hypothesis \citep{gilbert1986origin}, molecules are represented as sequences of numerical values. Analogous to how deep learning models process symbolic inputs such as texts or images, molecules are treated here as storage of information. Each molecule possesses a form of RNA duality, serving both as an information-storing entity and as a catalyst that facilitates the transformations of other molecules.

When molecules are in sufficient proximity, chemical reactions occur. These reactions are not modeled as simple bindings or exchanges but are assumed to implement non-linear transformations equivalent to a single layer of a neural network. Specifically, a catalyst molecule A encodes weights (and biases if needed) in its sequence, which it applies to an input molecule B to produce a new output molecule C. Transformation may result in a change in sequence length, corresponding to compression or expansion of information.

As such transformations repeat under environmental constraints, such as limits on the number of molecules or molecular lifetimes, competition and selection naturally emerge, leading to the formation of self-replicating networks. In such networks, molecules A and B may mutually catalyze each other's formation, creating a bidirectional reaction cycle that can be interpreted as an autoencoder structure.

If molecule B is shorter than molecule A, the transformation from A to B functions as encoding (compression), while the reverse transformation from B to A acts as decoding (reconstruction). This structure can be further extended hierarchically: B transforms into a shorter molecule C, C into D, and so forth. Through such iterative transformations, a multi-layered autoencoder network emerges. In this hierarchical structure, information is redundantly preserved across layers. Shallower layers correspond to lower-level physical features, while deeper layers represent more abstract or functional features. Mutations occurring at different layers can therefore lead to a range of evolutionary consequences, from minor physical variations to discontinuous shifts in functional meaning.

This model is not merely a conceptual illustration, but a concrete instantiation of how the hierarchical autoencoder mechanism proposed in this hypothesis can arise spontaneously. As shown in prior ALife research, such structured self-organization is indeed possible through the selection process. The present hypothesis holds that the spontaneous formation of these autoencoder molecular architectures is a key enabler of the large-scale and semantically meaningful evolutionary changes that often appear discontinuous or goal-directed.

\section{Simulating the Emergence of Stacked Autoencoding Structures}
\subsection{Simulation design and implementation}
The aim of the following simulation is not to provide an empirical verification of the Stacked Autoencoder Evolution Hypothesis, but rather to demonstrate that two of its core structural assumptions can arise even in a minimal model. Specifically, the simulation illustrates that (i) self replication can emerge as a closed loop of molecular transformations when the reaction space is constrained, and (ii) interactions among molecules of differing lengths can spontaneously produce layered relationships consistent with the hierarchical encoding structure proposed in the hypothesis. These observations are intended to clarify the conceptual plausibility of the mechanism, not to model biological evolution in detail.

It should be noted that the parameter settings and neural network architecture employed in this simulation are provided solely as illustrative examples to facilitate understanding of the hypothesis. These design choices carry no intrinsic significance, and the implementation presented here is merely one example. The simplicity of the model reflects its intended role as a conceptual demonstration rather than as a full evolutionary scenario.

In this simulation, a simplified chemical system was constructed using three types of molecular entities, each serving both as an information-storing sequence and as a catalyst for molecular transformations via neural network–based reactions. To focus on the core dynamics of the proposed mechanism, the model omits the complexity of a full evolutionary scenario involving diverse molecular populations. The three molecular types are denoted as $M_{13}$, $M_{7}$, and $M_{3}$, corresponding to sequences of lengths 13, 7, and 3, respectively. For the molecules of length 13, the $k$-th element of the $i$-th instance is represented as $M_{13, k}^{i}$, with each element $M_{\cdot, k}^{i} \in [-1, 1]$ initialized from a uniform distribution.

In this model, the ``state'' of a molecule is represented as a vector of real-valued components within the interval $[-1,1]$. These values do not correspond to any specific chemical property but serve as an abstract encoding of internal molecular degrees of freedom. The choice of the range $[-1,1]$ is made for mathematical consistency, as it matches the output of the hyperbolic tangent function used in the catalytic transformations. This continuous representation allows molecules to function as information-bearing entities whose states can be compressed, transformed, and reconstructed by the catalytic operators. These specific sequence lengths were chosen to allow the mapping of molecular sequences to neural network parameters. 

In addition to storing information, each molecule also functions as a catalyst that implements transformation processes \citep{ikegami1995active}. Conceptually, this means that a molecule of one type (e.g., $M_{13}$, $M_{7}$, or $M_{3}$) receives another molecule as input and produces a new molecule of one of the three types as output, thereby enabling the reaction network to support the mappings, $M_{13}\rightarrow M_{7}$, $M_{7}\rightarrow M_{3}$, $M_{7}\rightarrow M_{13}$, and so forth.

A straightforward way to implement such transformations would be to use a fully connected neural network, where the number of parameters is proportional to the product of the input and output dimensions. 
To avoid this issue, self-contained set of molecular species, the present model uses convolutional and deconvolutional neural networks \citep{lecun1998gradient,krizhevsky2012imagenet}. These architectures drastically reduce the number of parameters by reusing the same kernel across positions, while still allowing flexible control of output lengths through kernel size, padding, and stride. By choosing these parameters appropriately, the number of weight parameters required for each catalytic transformation can be kept aligned with the lengths of the three molecular types $M_{13}$, $M_{7}$, and $M_{3}$.

This design ensures that every catalytic molecule can encode exactly the parameters needed for its transformation function, thereby enabling a closed reaction cycle in which all transformations occur within a finite set of molecular species.

\begin{figure}[tp] 
    \centering
    \includegraphics[width=5.0in]{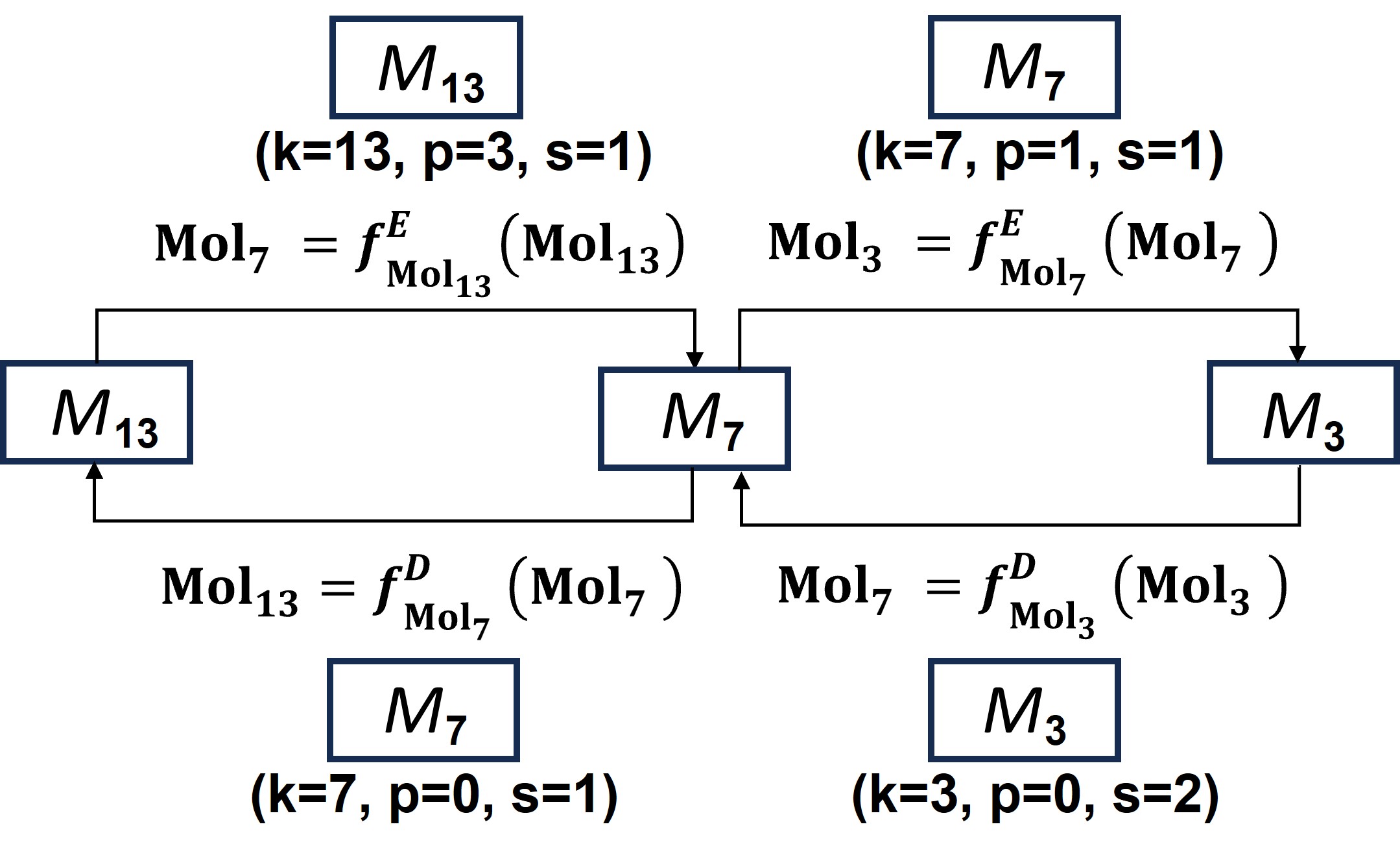}
    \caption{Artificial chemical reaction network in simulation. $M_{\#}$ represents molecules of length \#. Arrows indicate transformations from $M_{\#1}$ to $M_{\#2}$, where $M_{\#3}$ placed along the arrow serves as a catalyst. For example, the reaction in the upper left shows that a molecule $M_{13}^{i}$ is converted to a molecule $M_{7}^{j}$ through catalysis by another molecule $M_{13}^{l}$. Catalytic functions utilize convolutional networks (encoding) or deconvolutional networks (decoding). $f^{E}_{M_{\#1}}(M_{\#2})$ denotes a convolutional network constituted by $M_{\#1}$ that receives $M_{\#2}$ as input. The parameters $k$, $p$, and $s$ of the catalyst molecule represent kernel size, padding, and stride of convolutional or deconvolutional networks, respectively.}
    \label{fig:sim}
\end{figure}

The parameters and reaction designs used in the simulation are summarized in Fig.~\ref{fig:sim}. For the non-linear transformation within the catalytic reactions, a scaled hyperbolic tangent function $\tanh(2x)$ was used in order to expand the output range and facilitate the generation of larger variations. 
It should be emphasized again that this reaction network design represents just one possible example, and in real-world scenarios, the diversity of molecular species would allow for the construction of a wide variety of networks capable of forming loops in numerous ways.

At each simulation step, the model computes all possible catalytic reactions between molecules. For every ordered pair $(M_{X}, M_{Y})$ in the population, $M_{X}$ is treated as a substrate and $M_{Y}$ as a catalyst. The catalytic transformation encoded in $M_{Y}$ is applied to the state of $M_{X}$, generating a candidate product molecule $M_{Z}$. This procedure yields a large pool of potential reaction products.

From this pool, the next molecular population is constructed by randomly sampling new molecules. Specifically, ninety percent of the molecules in the next generation are sampled from the newly generated reaction products, and the remaining ten percent are copied from the current population to maintain continuity and prevent loss of diversity. The total population size is kept constant.

Because each molecule encodes the parameters of its catalytic transformation, any modification to a molecule directly alters the transformation it will impose in subsequent reactions. As a result, both molecular states and catalytic functions are continuously reshaped through the repeated application of reactions and perturbations. This recursive coupling produces a self-organizing dynamics in which molecular-level representations and reaction operators co-evolve.

The simulation environment consisted of fifteen thousand molecules, with five thousand molecules of each type $M_{13}$, $M_{7}$, and $M_{3}$. All molecules were initialized with random values, and no spatial structure was imposed. After each reaction cycle, small random perturbations were added to all molecules to promote variability. Through these cycles of reaction, sampling, and perturbation, the simulation investigates whether hierarchical autoencoding structures can emerge and stabilize within the molecular population. 

\subsection{Simulation results}

\begin{figure}[tp]
    \centering
    \includegraphics[width=7.0in]{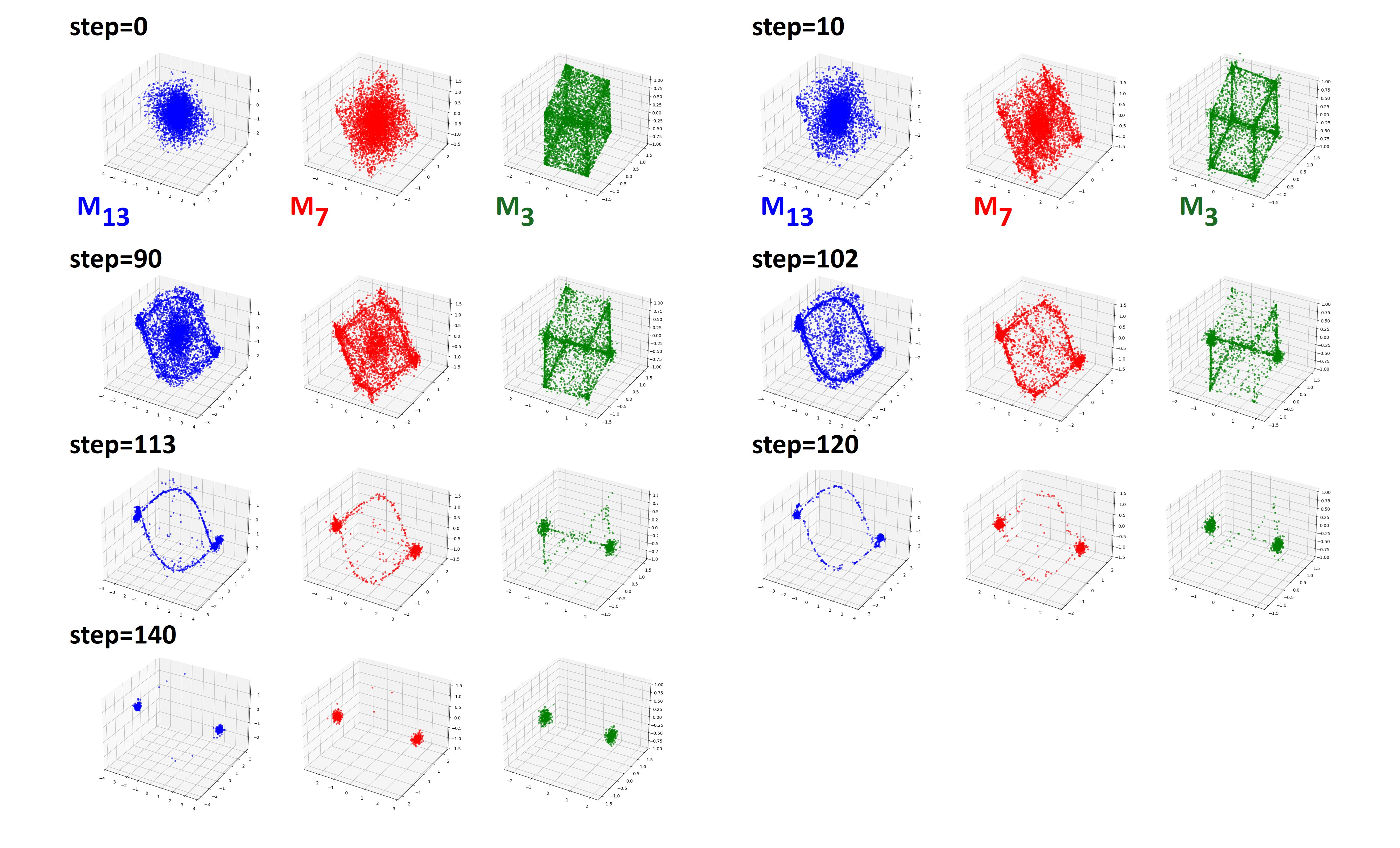}
    \caption{Distributions of $M_{13}$, $M_{7}$, and $M_{3}$ molecular states at each step. The distributions are visualized in three dimensions by applying principal component analysis (PCA) for dimensionality reduction.}
    \label{fig:particles}
\end{figure}

Based on the reaction network described above, the changes in the distribution of molecules of different lengths were examined through iterative reaction cycles. The results are shown in Fig.~\ref{fig:particles}. 

In the initial configuration, all molecular values were randomly initialized within the range from -1 to 1. After applying a single transformation step, no characteristic structural patterns were observed (shown as step = 0 in the figure).
As the reactions progress, however, a distinctive structure can be observed in the population of $M_{3}$ molecules. The cube-like distribution visible in the figure reflects the characteristic behavior of the hyperbolic tangent function, which tends to push values toward the extreme boundaries of its output range.
Although transient structural patterns occasionally appeared in the population of $M_{7}$ molecules until about 90 steps, seemingly influenced by the structure of $M_{3}$ molecules, these patterns did not persist. Over the long term, no clear structural organization was observed in either $M_{7}$ or $M_{13}$ populations.

Interestingly, at a certain point (around step 102 or 113 in Fig.~\ref{fig:particles}), the $M_{3}$ population began to converge into two distinct states. Subsequently, the $M_{7}$ and $M_{13}$ populations also underwent gradual changes, adopting slightly different structures while following a similar trend, and eventually, all molecule populations converged into two distinct clusters.

To evaluate whether the reaction network had acquired the ability to encode and decode information across molecular layers, we measured a reconstruction error based on mean squared error (MSE). The purpose of this measure is to assess how well the information in an original molecule is preserved after passing through a sequence of catalytic transformations. A low reconstruction error indicates that the intermediate layers retain enough information for the original molecule to be reliably reconstructed, analogous to the behaviour of a stacked autoencoder.

In general, the evaluation proceeds by (i) taking a molecule of one type, (ii) transforming it through one or more intermediate molecular layers, and (iii) comparing the final reconstructed molecule with members of the original population of the same type. This allows us to quantify the extent to which information survives through the full encode–decode pathway.

As an example, the reconstruction accuracy for $M_{13}$ molecules was assessed as follows. Two molecules of $M_{13}$, denoted $M_{13}^{(i)}$ and $M_{13}^{(j)}$, were reacted to produce a molecule of $M_{7}$, denoted $M_{7}^{(k)}$. Subsequently, two such $M_{7}$ molecules, $M_{7}^{(k)}$ and $M_{7}^{(l)}$, were reacted to generate a new $M_{13}$ molecule, denoted $M_{13}^{(new)}$. For each reconstructed molecule, the MSE was computed with respect to its closest match in the original $M_{13}$ population, yielding the reconstruction error

\begin{equation}
RE_{13} = \frac{1}{5000} \sum_{i=1}^{5000} \min_{j} \mathrm{MSE}(M_{13}^{(new,i)}, M_{13}^{(orig,j)}).
\end{equation}

This procedure quantifies reconstruction through the pathway 
$M_{13} \rightarrow M_{7} \rightarrow M_{13}$.  
The same method was applied to other pathways, including 
$M_{7} \rightarrow M_{3} \rightarrow M_{7}$ and 
$M_{13} \rightarrow M_{7} \rightarrow M_{3} \rightarrow M_{7} \rightarrow M_{13}$,
allowing reconstruction performance to be evaluated across multiple hierarchical transitions.

\begin{figure}[tp]
    \centering
    \includegraphics[width=5.0in]{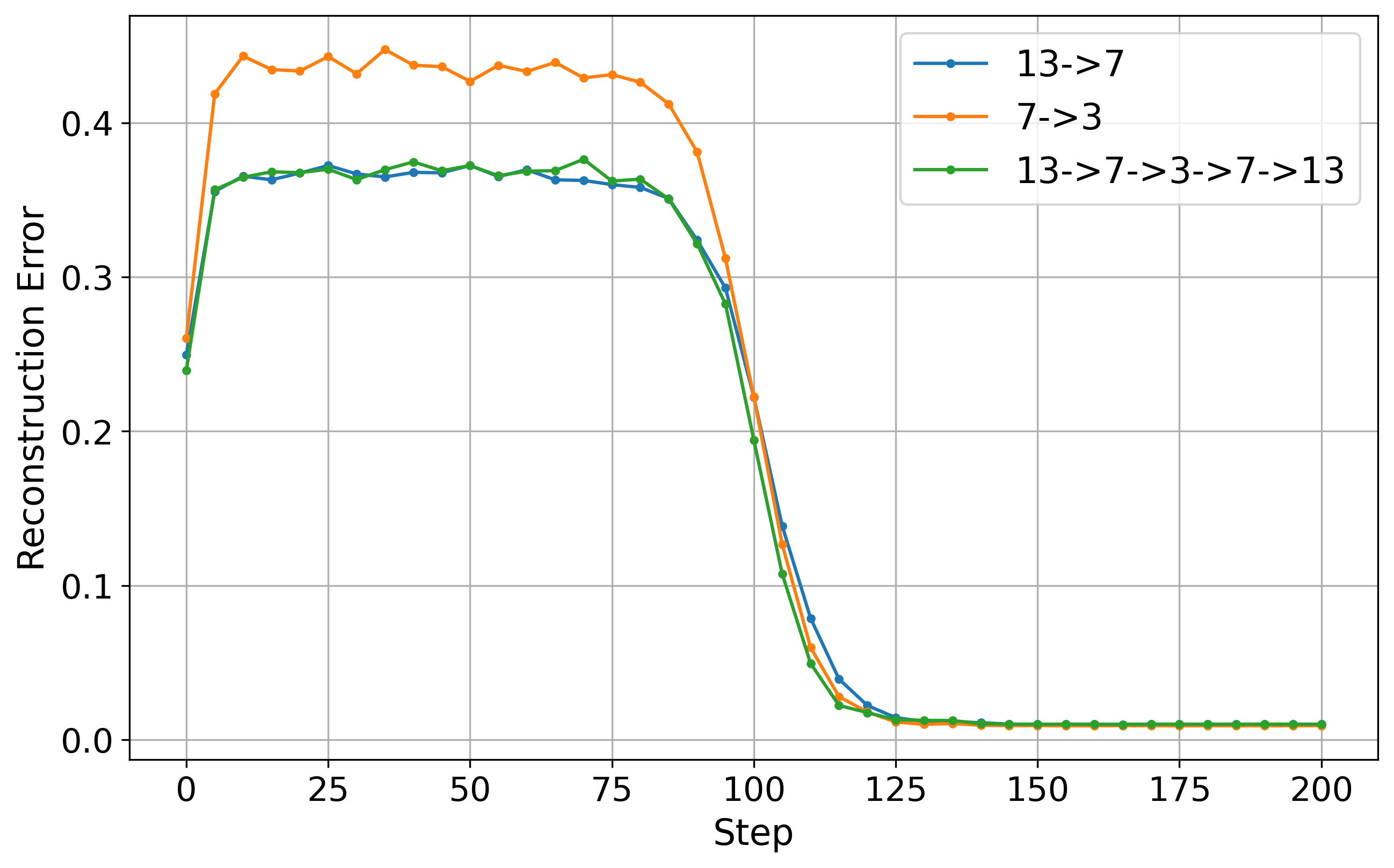}
    \caption{Reconstruction errors at each step. See main text for details.}
    \label{fig:rec_errors}
\end{figure}

Figure~\ref{fig:rec_errors} shows the changes in the reconstruction errors over time. Initially, the reconstruction error was very large, but as the system approached the two-state convergence, the error rapidly decreased. In the final two-state configuration, it is reasonable to assume that all layers retain the same information, as evidenced by the reduction in reconstruction error.

\begin{figure}[tp]
    \centering
    \includegraphics[width=5.0in]{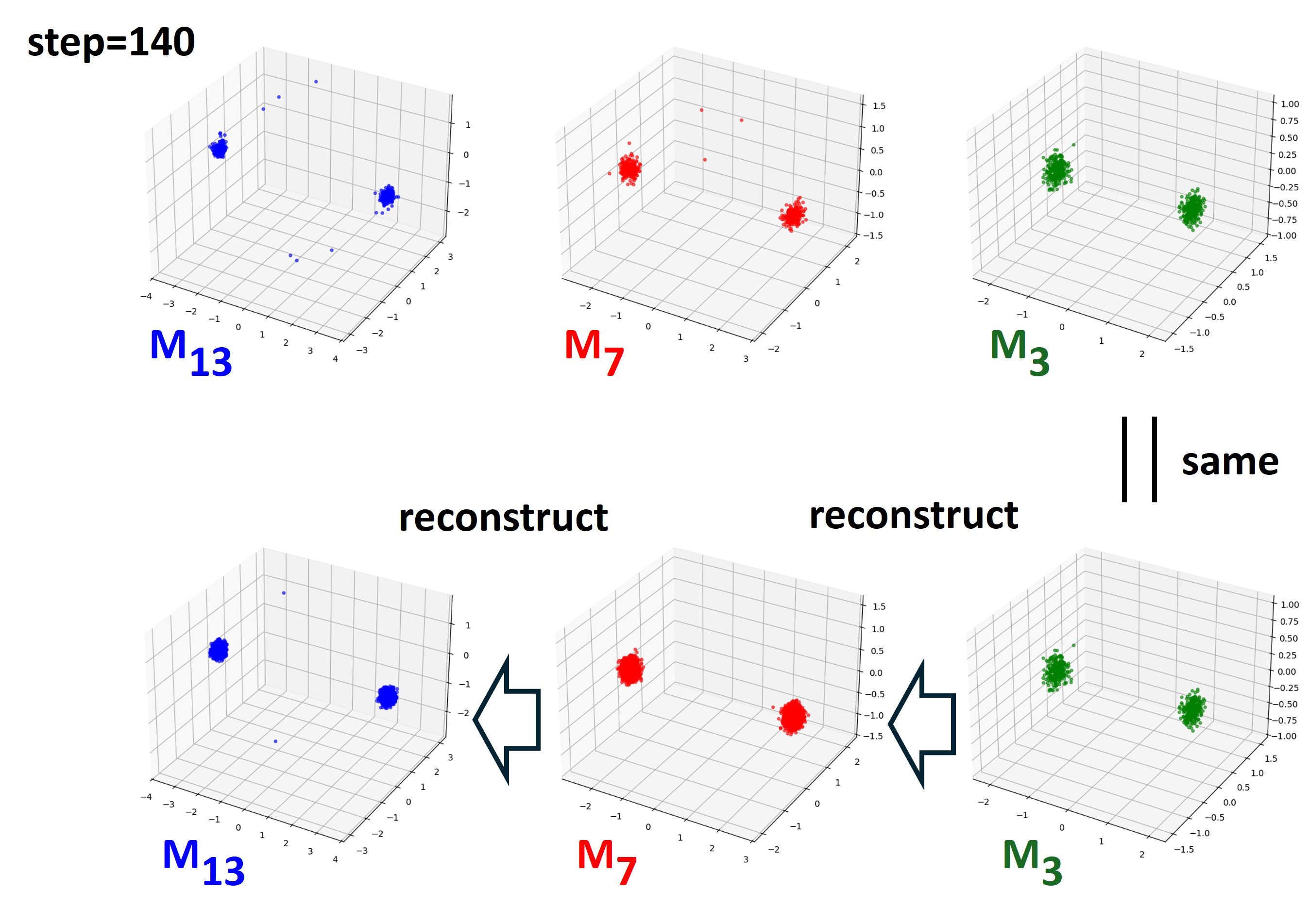}
    \caption{Results of decoding $M_{7}$ and $M_{13}$ molecules from $M_{3}$ molecules. The top panels show the original distributions at step 129, while the bottom panels present the distributions obtained by decoding from the same $M_{3}$ molecules.}
    \label{fig:reconstruction}
\end{figure}

To examine whether the $M_{3}$ molecules retained information about the $M_{7}$ and $M_{13}$ molecules, we reconstructed $M_{7}$ and $M_{13}$ from the $M_{3}$ molecules using the state at step 140. The results are shown in Fig.~\ref{fig:reconstruction}. As shown in the figure, both $M_{7}$ and $M_{13}$ molecular state distributions were successfully reconstructed from the $M_{3}$ molecules. This suggests that the information necessary to specify the states of the lower layers was embedded in the higher-level representations of the $M_{3}$ molecules.

These results suggest that when a self-catalytic reaction network is subjected to selection pressure for self-replication, it can spontaneously develop a hierarchical information encoding and decoding structure that was proposed in this paper.

\subsection{Limitations of the model}
In the present simulation, the selection pressure is extremely simple, focusing solely on the ease with which numerical sequences can be replicated. As a result, there is no driving force that promotes an increase in sequence complexity. Consequently, all layers eventually converge to binary states. Under such simplified conditions, there is no inherent need for the sequences to become more complex, nor is there a necessity for information compression between layers. This outcome reflects the overly simplistic nature of the selection pressure defined in this model.

In more realistic evolutionary settings, the situation is fundamentally different. Although replication and the resulting competition for limited spatial or material resources remain the primary source of selection, these basic pressures manifest indirectly as a wide range of survival- and robustness-related constraints. For example, organisms must maintain structural integrity, withstand environmental fluctuations, and coordinate multiple functional processes in order to continue occupying ecological space. Such indirect consequences of the underlying competition can generate pressures that favour increased molecular or developmental complexity. Once sequences become sufficiently complex under these conditions, the hierarchical information compression and redundant multi-layer representations proposed in this study may become more pronounced and more readily observable.

It is important to emphasize that the present model is intended as a conceptual demonstration. It abstracts away biochemical constraints, physical embodiment, and population-level evolutionary dynamics. Moreover, the hypothesis currently lacks direct biological evidence, and identifying molecular realizations of the proposed multi-layered encoding architecture remains an open challenge. Nevertheless, as is common in theoretical artificial-life research, the value of the framework lies in demonstrating a constructive possibility: that hierarchical encoding and decoding can emerge within a self-replicating system and may, in principle, support discontinuous evolutionary transitions. By clarifying both the generative potential and the limitations of this mechanism, the hypothesis provides a foundation for future empirical, computational, and synthetic investigations into multi-stage information processing in evolution.

\section{Discussion with Other Evolution-as-Learning Hypotheses}
The relationship between evolution and learning has long been explored in evolutionary biology, with various conceptual frameworks proposed. Among these, the Baldwin effect remains a particularly influential concept \citep{baldwin1896new}. The Baldwin effect describes a mechanism in which learning interacts with genetic evolution, positioning learning itself as a target of natural selection. Organisms capable of adapting through learning gain a survival advantage, and over evolutionary timescales the neural and developmental architectures that support such learning become selectively stabilized. Furthermore, it has been suggested that behaviors or traits initially acquired through learning may eventually become genetically assimilated, allowing them to emerge without learning through genetic fixation.

This conceptual mechanism has been formalized through mathematical modeling and simulations by Hinton and Nowlan, who demonstrated that learning can significantly accelerate evolutionary processes \citep{hinton1987learning}. Building on this foundation, Watson and Szathmáry proposed that evolution itself can be understood as a learning process in which the evolutionary landscape is progressively ``memorized'' and shaped by evolutionary history \citep{watson2016evolution}. In their view, this memory is encoded not only in correlations among genes and modular developmental mechanisms, but also in the ecological relationships that organisms construct and modify through their interactions. This perspective treats evolving systems as inherently self-modifying, capable of reorganizing both their internal developmental structure and key aspects of the selective environment. Their framework assumes the existence of a fitness function that evaluates the ``goodness'' of states, allowing them to frame evolution as an optimization process analogous to learning, even though this fitness function is not necessarily fixed or absolute.

Vanchurin and others later extended this line of thought by suggesting that evolution proceeds not only by fitness optimization but also by the minimization of prediction error with respect to the environment \citep{vanchurin2022toward}. In this view, selection favors mechanisms that improve predictive or anticipatory behaviour. Both \citet{watson2016evolution} and \citet{vanchurin2022toward} emphasize that evolutionary systems operate across multiple levels of organization, giving rise to hierarchical adaptive dynamics that resemble learning across different scales.

The learning process implied by the present hypothesis differs from these approaches in its focus. Rather than interpreting evolution as an optimization or prediction process at the population level, the stacked–autoencoder hypothesis proposes that learning-like dynamics may already be embedded within the self-replication process itself. Here, self-replication is not treated as a direct duplication of genetic information but as a sequence of transformations that compress and reconstruct this information across multiple layers of representation, analogous to hierarchical autoencoders in deep learning. This process is hypothesized to give rise to multi-level latent spaces that encode increasingly abstract (or semantic) features of genetic information \citep{hinton2006reducing, bengio2013representation}. Mutations at different layers perturb different levels of abstraction, allowing both continuous and discontinuous evolutionary changes through the decoder process.

This framework does not replace fitness-based selection or the learning-like dynamics proposed by Hinton, Watson, or Vanchurin, nor does it conflict with more conventional evolutionary theories. Rather, it complements these perspectives by suggesting that the informational architecture of self-replication itself may endow evolutionary systems with a form of hierarchical representation learning. In doing so, the hypothesis offers a generative mechanism that can link microscopic perturbations in latent representations to macroscopic evolutionary transitions, thereby providing one possible route by which evolution can exhibit both continuity and discontinuity within a unified informational process.

\section{Conclusion}
This study has proposed a novel theoretical framework for evolution, the Stacked Autoencoder Evolution Hypothesis. This hypothesis suggests that evolution is not merely a process of self-replication, but a dynamic process of hierarchical information transformation in which genetic encoding is compressed and reconstructed across multiple layers of abstraction. As a result, evolution may not operate solely through superficial sequence-level mutations, but through perturbations in abstract semantic spaces, enabling rapid and adaptive evolutionary shifts.

By introducing the computational principle of autoencoder into evolutionary theory, this hypothesis offers a new computational perspective on the evolutionary process. It suggests the potential to bridge traditional evolution theory with advanced computational frameworks such as self-supervised learning and representation learning, which have been actively developed in contemporary machine learning. For instance, it opens up the possibility that the advanced neural architectures, such as the Transformer models that have achieved remarkable success in natural language processing, might be inherently embedded in evolutionary processes as well. Just as AI systems are capable of generating diverse and meaningful linguistic outputs, evolution itself may have acquired the capacity to operate within a semantic space as rich and structured as human language.

In contrast to descriptive accounts such as punctuated equilibrium or saltation, which characterise the patterns of rapid change, the present framework specifies a concrete mechanism by which coordinated, system-wide novelties can be generated. In particular, mutations at deeper latent layers perturb compressed representations that, when decoded, induce structured changes across many traits simultaneously, thereby linking microscopic informational perturbations to macroscopic evolutionary transitions within a single generative process.

This hypothesis reframes evolution as an emergent computational process rooted in hierarchical information transformation. By moving beyond the limitations of random mutation and selection, it offers a generative account of how large-scale and semantically structured evolutionary shifts can arise from the informational architecture of genetic encoding.

\section*{Code Availability}
The code used for the simulations and analyses in this study is publicly available at:\\
\texttt{https://github.com/HiroIizuka/SAEvo}.

\printbibliography

@article{hinton2006reducing,
  title={Reducing the dimensionality of data with neural networks},
  author={Hinton, Geoffrey E and Salakhutdinov, Ruslan R},
  journal={science},
  volume={313},
  number={5786},
  pages={504--507},
  year={2006},
  publisher={American Association for the Advancement of Science}
}

@article{asquith2007sources,
  title={Sources for Imanishi Kinji’s views of sociality and evolutionary outcomes},
  author={Asquith, Pamela J},
  journal={Journal of Biosciences},
  volume={32},
  number={4},
  pages={635--641},
  year={2007},
  publisher={Springer}
}

@article{kimura1979neutral,
  title={The Neutral Theory},
  author={Kimura, Motoo},
  journal={Scientific American},
  year={1979}
}

@book{darwin1859,
  title={On the origin of species, 1859},
  author={Darwin, Charles},
  year={1859},
  publisher={London: John Murray}
}

@article{lecun1998gradient,
  title={Gradient-based learning applied to document recognition},
  author={LeCun, Yann and Bottou, L{\'e}on and Bengio, Yoshua and Haffner, Patrick},
  journal={Proceedings of the IEEE},
  volume={86},
  number={11},
  pages={2278--2324},
  year={1998},
  publisher={IEEE}
}

@article{mcmenamin2013cambrian,
  title={The Cambrian Explosion: The Construction of Animal Biodiversity.},
  author={McMenamin, Mark AS},
  journal={BioScience},
  volume={63},
  number={10},
  pages={834--835},
  year={2013},
  publisher={BioOne}
}

@book{valentine2004origin,
  title={On the origin of phyla},
  author={Valentine, James W},
  year={2004},
  publisher={University of Chicago Press}
}

@book{carroll2005endless,
  title={Endless forms most beautiful: The new science of {Evo}-{Devo} and the making of the animal kingdom},
  author={Carroll, Sean B},
  number={54},
  year={2005},
  publisher={WW Norton \& Company}
}

@book{schluter2000ecology,
  title={The ecology of adaptive radiation},
  author={Schluter, Dolph},
  year={2000},
  publisher={OUP Oxford}
}

@article{tria2014dynamics,
  title={The dynamics of correlated novelties},
  author={Tria, Francesca and Loreto, Vittorio and Servedio, Vito Domenico Pietro and Strogatz, Steven H},
  journal={Scientific reports},
  volume={4},
  number={1},
  pages={5890},
  year={2014},
  publisher={Nature Publishing Group UK London}
}

@book{kauffman2000investigations,
  title={Investigations},
  author={Kauffman, Stuart A},
  year={2000},
  publisher={Oxford University Press}
}

@article{watson2016evolution,
  title={How can evolution learn?},
  author={Watson, Richard A and Szathm{\'a}ry, E{\"o}rs},
  journal={Trends in ecology \& evolution},
  volume={31},
  number={2},
  pages={147--157},
  year={2016},
  publisher={Elsevier}
}

@article{krizhevsky2012imagenet,
  title={Imagenet classification with deep convolutional neural networks},
  author={Krizhevsky, Alex and Sutskever, Ilya and Hinton, Geoffrey E},
  journal={Advances in neural information processing systems},
  volume={25},
  year={2012}
}

@article{bengio2013representation,
  title={Representation learning: A review and new perspectives},
  author={Bengio, Yoshua and Courville, Aaron and Vincent, Pascal},
  journal={IEEE transactions on pattern analysis and machine intelligence},
  volume={35},
  number={8},
  pages={1798--1828},
  year={2013},
  publisher={IEEE}
}

@inproceedings{erhan2010does,
  title={Why does unsupervised pre-training help deep learning?},
  author={Erhan, Dumitru and Courville, Aaron and Bengio, Yoshua and Vincent, Pascal},
  booktitle={Proceedings of the thirteenth international conference on artificial intelligence and statistics},
  pages={201--208},
  year={2010},
  organization={JMLR Workshop and Conference Proceedings}
}

@article{bengio2006greedy,
  title={Greedy layer-wise training of deep networks},
  author={Bengio, Yoshua and Lamblin, Pascal and Popovici, Dan and Larochelle, Hugo},
  journal={Advances in neural information processing systems},
  volume={19},
  year={2006}
}

@article{vincent2010stacked,
  title={Stacked denoising autoencoders: Learning useful representations in a deep network with a local denoising criterion.},
  author={Vincent, Pascal and Larochelle, Hugo and Lajoie, Isabelle and Bengio, Yoshua and Manzagol, Pierre-Antoine and Bottou, L{\'e}on},
  journal={Journal of machine learning research},
  volume={11},
  number={12},
  year={2010}
}

@inproceedings{masci2011stacked,
  title={Stacked convolutional auto-encoders for hierarchical feature extraction},
  author={Masci, Jonathan and Meier, Ueli and Cire{\c{s}}an, Dan and Schmidhuber, J{\"u}rgen},
  booktitle={Artificial neural networks and machine learning--ICANN 2011: 21st international conference on artificial neural networks, espoo, Finland, June 14-17, 2011, proceedings, part i 21},
  pages={52--59},
  year={2011},
  organization={Springer}
}

@inproceedings{ngiam2011multimodal,
  title={Multimodal deep learning.},
  author={Ngiam, Jiquan and Khosla, Aditya and Kim, Mingyu and Nam, Juhan and Lee, Honglak and Ng, Andrew Y and others},
  booktitle={ICML},
  volume={11},
  pages={689--696},
  year={2011}
}

@article{hickinbotham2016maximizing,
  title={Maximizing the adjacent possible in automata chemistries},
  author={Hickinbotham, Simon and Clark, Edward and Nellis, Adam and Stepney, Susan and Clarke, Tim and Young, Peter},
  journal={Artificial Life},
  volume={22},
  number={1},
  pages={49--75},
  year={2016},
  publisher={MIT Press}
}

@article{ikegami1995active,
  title={Active mutation in self-reproducing networks of machines and tapes},
  author={Ikegami, Takashi and Hashimoto, Takashi},
  journal={Artificial Life},
  volume={2},
  number={3},
  pages={305--318},
  year={1995}
}

@Book{varela1979,
  author =	 {Varela, Francisco J.},
  title =	 {Principles of Biological Autonomy},
  publisher =	 {Elsevier North Holland, New York, NY},
  year =	 1979
}

@article{langton1984self,
  title={Self-reproduction in cellular automata},
  author={Langton, Christopher G},
  journal={Physica D: Nonlinear Phenomena},
  volume={10},
  number={1-2},
  pages={135--144},
  year={1984},
  publisher={Elsevier}
}

@article{ray1992,
  title={An approach to the synthesis of life},
  author={Ray, T. S.},
  journal={Proceedings of Artifcial Life},
  volume={II},
  pages={371--408},
  year={1992},
}

@article{ofria2004avida,
  title={Avida: A software platform for research in computational evolutionary biology},
  author={Ofria, Charles and Wilke, Claus O},
  journal={Artificial life},
  volume={10},
  number={2},
  pages={191--229},
  year={2004},
  publisher={MIT Press}
}

@article{sayama1999new,
  title={A new structurally dissolvable self-reproducing loop evolving in a simple cellular automata space},
  author={Sayama, Hiroki},
  journal={Artificial Life},
  volume={5},
  number={4},
  pages={343--365},
  year={1999},
  publisher={MIT Press}
}

@article{salzberg2003emergent,
  title={Emergent evolutionary dynamics of self-reproducing cellular automata},
  author={Salzberg, Christopher},
  journal={Master's thesis, Universiteit van Amsterdam, Amsterdam, The Netherlands},
  year={2003}
}

@article{gilbert1986origin,
  title={Origin of life: The {RNA} world},
  author={Gilbert, Walter},
  journal={nature},
  volume={319},
  number={6055},
  pages={618--618},
  year={1986},
  publisher={Nature Publishing Group UK London}
}

@article{vanchurin2022toward,
  title={Toward a theory of evolution as multilevel learning},
  author={Vanchurin, Vitaly and Wolf, Yuri I and Katsnelson, Mikhail I and Koonin, Eugene V},
  journal={Proceedings of the National Academy of Sciences},
  volume={119},
  number={6},
  pages={e2120037119},
  year={2022},
  publisher={National Academy of Sciences}
}

@article{hinton1987learning,
  title={How learning can guide evolution},
  author={Hinton, Geoffrey E and Nowlan, Steven},
  journal={Complex systems},
  volume={1},
  number={3},
  pages={495--502},
  year={1987}
}

@article{baldwin1896new,
  title={A new factor in evolution},
  author={Baldwin, J Mark},
  journal={The American Naturalist},
  volume={30},
  pages={441--451},
  year={1896},
}

@article{eldredge1972punctuated,
  title={Punctuated equilibria: an alternative to phyletic gradualism},
  author={Eldredge, Niles and Gould, Stephen Jay and others},
  journal={Models in paleobiology},
  volume={82},
  pages={115},
  year={1972},
  publisher={San Francisco}
}

@book{goldschmidt1940,
  author    = {Goldschmidt, Richard},
  title     = {The Material Basis of Evolution},
  year      = {1940},
  publisher = {(Repreinted by Yale University Press, 1982},
}

\end{document}